\documentclass[conference]{IEEEtran}
\IEEEoverridecommandlockouts
\usepackage{cite}
\usepackage{amsmath,amssymb,amsfonts}
\usepackage{algorithmic}
\usepackage{graphicx}
\usepackage{float}
\usepackage{subfigure}
\usepackage{textcomp}
\usepackage{xcolor}
\usepackage{multirow}
\usepackage{tikz}
\newcommand{\blackcircled}[1]{\tikz[baseline=(char.base)]{
    \node[shape=circle,draw,fill=black,inner sep=0.5pt] (char) {\textcolor{white}{#1}};}}
\newcommand{\blacksquared}[1]{\tikz[baseline=(char.base)]{
    \node[shape=rectangle,draw,fill=black,inner sep=2pt] (char) {\textcolor{white}{#1}};}}
\usepackage{siunitx}

\def\BibTeX{{\rm B\kern-.05em{\sc i\kern-.025em b}\kern-.08em
    T\kern-.1667em\lower.7ex\hbox{E}\kern-.125emX}}
\begin{document}

\bstctlcite{IEEEexample:BSTcontrol}
\title{An Efficient Data Reuse with Tile-Based Adaptive Stationary for Transformer Accelerators
}
\author{\IEEEauthorblockN{Tseng-Jen Li, and Tian-Sheuan Chang}

\IEEEauthorblockA{\textit{{Institute of Electronics, National Yang Ming Chiao Tung University}}, Hsinchu 300093, Taiwan
}
\IEEEauthorblockA{\textit{ericlee329@gmail.com, tschang@nycu.edu.tw}}
}

\maketitle

\begin{abstract}
Transformer-based models have become the \textit{de facto} backbone across many fields, such as computer vision and natural language processing. However, as these models scale in size, external memory access (EMA) for weight and activations becomes a critical bottleneck due to its significantly higher energy consumption compared to internal computations. While most prior work has focused on optimizing the self-attention mechanism, little attention has been given to optimizing data transfer during linear projections, where EMA costs are equally important. In this paper, we propose the Tile-based Adaptive Stationary (TAS) scheme that selects the input or weight stationary in a tile granularity, based on the input sequence length. Our experimental results demonstrate that TAS can significantly reduce EMA by more than 97\% compared to traditional stationary schemes, while being compatible with various attention optimization techniques and hardware accelerators.
\end{abstract}

\section{Introduction}

Transformer-based models \cite{vaswani2017attention} are now crucial in deep learning for a range of tasks. They process tokenized text, audio, or images with multiple transformer layers, excelling in handling long-range dependencies and parallel processing with attention mechanisms. Models like Vision Transformer (ViT) \cite{dosovitskiy2020image}, BERT \cite{devlin2018bert}, and Wav2Vec2.0 \cite{baevski2020wav2vec} have greatly outperformed older convolutional and recurrent models in accuracy. Their success is largely due to parameter scaling, where increased parameters through deeper layers, more attention heads, or larger embeddings improve performance. This strategy has led to large models like ViT-G/14 \cite{zhai2022scaling}, Wav2Vec2-XLS-R-2B \cite{babu2021xls}, and GPT-3 \cite{brown2020language}, as shown in Table~\ref{tab:EMA_diff_models}, which consistently push the accuracy limits. However, the growing number of parameters presents challenges for hardware accelerators, requiring efficient data reuse to reduce memory bandwidth constraints.

Data reuse through different stationary schemes has been well explored for accelerators of convolutional neural networks~\cite{chen2016eyeriss}. \cite{chen2016eyeriss} proposed the input/weight/output stationary schemes that reuse input/weight/output, respectively, for the computing flow. However, it is optimized for convolution-based computation. These stationary schemes have been applied in \cite{qin2024ayaka} that propose an accelerator for versatile transformer-based models. However, their stationary schemes are fixed when focusing solely on one of the transformer models. This approach is not optimal, since transformer input length could vary dramatically, such as the text or audio input. Using the BERT model as a reference, by inputting the tokenized text of length 3072 into the adaptive stationary strategy, it is possible to obtain over a 75\% reduction in the reused matrix as compared to \cite{qin2024ayaka}, which shows that an optimal stationary scheme shall be input-length dependent. Moreover, their method necessitates concurrent read and write operations to external memory because of data flow conflicts, which imposes further stall penalties on the hardware.

To mitigate these challenges, we propose the \textit{Tile-based Adaptive Stationary} (TAS) that selects either input or weight stationary according to input length to maximize data reuse, reduce external memory access, as well as decrease the amount of simultaneous reading from and writing to external memory access. TAS can reduce external memory accesses by approximately 97\% compared to the one without any data reuse. 
Besides, this approach optimizes memory communication by achieving nearly twice the efficiency compared to the previous fixed stationary method~\cite{qin2024ayaka}.

The rest of the paper is organized as follows. Section II reviews fixed stationary methods and its problems. Section III presents our proposed methods. Section IV shows the experimental results. Finally, this paper is concluded in Section V.

\begin{table*}[t]
        \begin{minipage}{0.4\linewidth} %
            \begin{center}
                \caption{Hidden Dimension, Pre-defined Token Length, Parameter Size and Total EMA Values for Respective Representative Large Model}
                \label{tab:EMA_diff_models}
                \begin{tabular}{|c|c|c|c|} \hline
                     Model & ViT-G/14 &  \shortstack{Wav2Vec2 \\ -XLS-R} &  GPT-3 \\ \hline
                     \shortstack{Hidden \\ Dimension} & 4,096 & 2,560 & 12,288 \\ \hline
                     \shortstack{Token \\ Length} & 518 & 1,536 & 2,048 \\ \hline
                     \shortstack{Parameter \\ Size (B)} & 1.8 & 2 & 175 \\ \hline
                     \shortstack{Total \\ EMA (G)} & 312.9 & 353.9 & 11,132.6 \\ \hline
                \end{tabular}                                
            \end{center}
        \end{minipage}%
        \begin{minipage}{0.6\linewidth} %
            \begin{center}
                \renewcommand{\arraystretch}{1.5} %
                \caption{Comparison of External Memory Access for Various Stationary Schemes}
                \label{tab:ema_for_stationary_schemes}
                \begin{tabular}{|c|c|c|c|c|}
                    \hline
                    \shortstack{\textbf{Stationary} \\ \textbf{Scheme}} & 
                    \shortstack{\textbf{Input} \\ \textbf{Matrix}} & 
                    \shortstack{\textbf{Weight} \\ \textbf{Matrix}} & 
                    \shortstack{\textbf{Output} \\ \textbf{Matrix}} & 
                    \textbf{Total} \\ \hline
                    \textbf{Naïve} & \(K \times MN\) & \(M \times NK\) & \(N \times MK\) & \(MNK \times 3\) \\ \hline
                    \textbf{IS} & \(\textbf{MN}\) & \(\frac{M}{m} \times NK\) & \(\frac{N}{n} \times MK\) &
                    \(MNK\times(\frac{1}{K}+\frac{1}{m}+\frac{1}{n})\) \\ \hline
                    \textbf{WS} & \(\frac{K}{k} \times MN\) & \(\textbf{NK}\) & \(\frac{N}{n} \times MK\) &
                    \(MNK\times(\frac{1}{k}+\frac{1}{M}+\frac{1}{n})\) \\ \hline
                    \textbf{OS} & \(\frac{K}{k} \times MN\) & \(\frac{M}{m} \times NK\) & \(MK\) &
                    \(MNK\times(\frac{1}{k}+\frac{1}{m}+\frac{1}{N})\) \\ \hline
                    \textbf{IS-OS} & \(MN\) & \(\frac{M}{m} \times NK\) & \(MK\) & 
                    \(MNK\times(\frac{1}{K}+\frac{1}{m}+\frac{1}{N})\) \\ \hline
                    \textbf{WS-OS} & \(\frac{K}{k} \times MN\) & \(NK\) & \(MK\) & 
                    \(MNK\times(\frac{1}{k}+\frac{1}{M}+\frac{1}{N})\) \\ \hline
                \end{tabular}                
            \end{center}
        \end{minipage}%
    \end{table*} 

\begin{figure}
    \centering
    \includegraphics[width=1\linewidth]{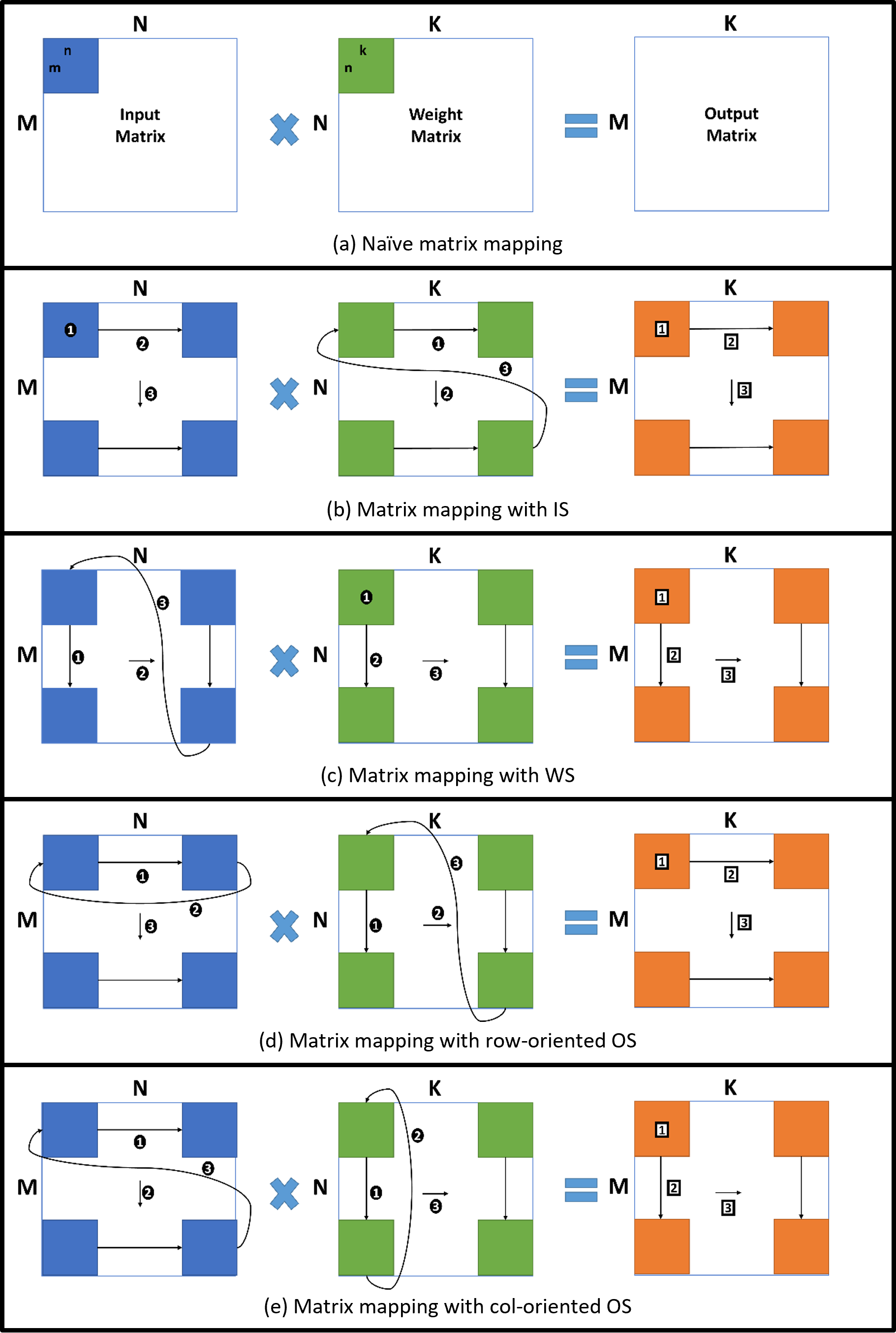}
    \caption{Matrix Mapping for Matrix-Matrix Multiplication with Conventional Stationary Schemes}
    \label{fig:CSS}
\end{figure}

\section{Review of the Fixed Stationary Scheme}

Fig.~\ref{fig:CSS} shows the fixed stationary strategies for matrix-matrix multiplication, often employed in transformer models for operations like linear projections. Though these stationary strategies are similar to those in CNNs, they feature unique data flows and memory access patterns, detailed in this Section. Details for Fig.~\ref{fig:CSS} are as follows. In Fig.~\ref{fig:CSS}(a), $M$ is the input matrix row count, $K$ is the weight matrix column count, and $N$ is the shared dimension between input matrix columns and weight matrix rows. The parameters $m$, $n$, and $k$ denote the tile size, where $m \leq M$, $n \leq N$, and $k \leq K$. Circled numbers indicate tile computation order. Arrows with numbers show tile advancement direction post-computation, whereas numbers without arrows suggest tiles remain until aligned with partners. Squared numbers signify the sequence of output results. The different stationary schemes are outlined below.

\paragraph{Input Stationary}

The \textbf{IS} scheme aims to load input tile data into internal memory once for reuse, reducing memory access to retrieve identical inputs. Fig.~\ref{fig:CSS}(b) portrays data movement in the \textbf{IS} scheme. An arrow labeled \blackcircled{1} shows weight tile data progressing along weight matrix rows, denoted $K$. Thus, the input tile marked \blackcircled{1} is used $K/k$ times. The \blackcircled{2} arrow indicates that the input tile in the input matrix moves $n$ units, while the weight tile shifts $n$ units downward in the weight matrix. As the input tile transitions from west to east in the input matrix, it moves $m$ units to the next row, repositioning the weight tile from right-bottom to left-top in the weight matrix. An arrow marked \blacksquared{2} produces the output tile. These steps repeat $M/m$ times, as indicated by the arrow marked \blackcircled{3}.

\paragraph{Weight Stationary}

The \textbf{WS} scheme loads each weight tile into internal memory once, enabling repeated use across multiple calculations, which reduces redundant memory access. Fig.~\ref{fig:CSS}(c) illustrates this dataflow. An arrow \blackcircled{1} indicates the fixed position of the weight tile along matrix columns, $M$, where each tile is reused $M/m$ times. Arrow \blackcircled{2} shows the input tile moving $k$ units down the input matrix, while the stationary weight tile remains fixed. As the input tile traverses from east to west, the weight tile remains in place and is reused for subsequent inputs, shifting vertically by $k$. The output tile generation, marked by \blacksquared{2}, follows this sequence and repeats $K/k$ times, indicated by \blackcircled{3}.

\paragraph{Output Stationary}
The \textbf{OS} strategies depicted in Fig.~\ref{fig:CSS}(d) and (e) showcase another commonly employed group of techniques. These approaches retain partial sums in internal memory until the computation of output results is completed, thereby optimizing inner products, which are fundamental to matrix-matrix multiplication. This method leverages the spatial locality of processing element arrays to reduce external data retrievals for partial sums. The primary distinction between the \textbf{row-oriented OS} and \textbf{col-oriented OS} schemes lies in the sequence of generating the weight matrix. In the row-oriented OS, outputs are generated row by row, while the column-oriented strategy produces results column by column. Fig.~\ref{fig:CSS}(d) shows arrows with black circle indicators where the input and weight tiles slide vertically and horizontally, respectively. Upon generating an output tile marked by \blacksquared{1}, indicated by arrows in \blackcircled{2}, the input tile returns to the initial position in the input matrix, and the weight tile shifts right by $k$. After repeating the data flow $K/k$ times, as shown by arrows marked as \blackcircled{3}, the input tile is moved downward by $m$, and the weight tile is moved to the starting weight matrix location. On the other hand, the column-oriented strategy initially redirects the weight tile, as shown by arrows in \blackcircled{2}, and later shifts the input tile, indicated by \blackcircled{3}.  

\paragraph{Problems of the fixed stationary scheme}
Table~\ref{tab:ema_for_stationary_schemes} shows the external memory access values for different stationary schemes. These fixed stationary schemes have two problems: higher external memory access and concurrent read/write demands. For the first problem, in this table, $m$, $n$, and $k$ represent tile dimensions and the available hardware computation resources, typically much smaller than $M$, $N$, and $K$. For instance, with an IS configuration, the EMA of the input matrix is noticeably less compared to others, highlighting the overall EMA reduction. Therefore, applying a fixed stationary scheme without considering the input length is inadvisable. For the second problem, external memory like DRAM cannot read and write data simultaneously. However, when computing linear projections, two matrices need to be read from external memory, while only one requires writing back. Though this can solved by extra buffers, the required buffer size could be quite large due to the large matrix size in current transformer models. Thus, an effective partial sum reuse scheme is demanded to solve this problem.

\section{Methodology}
This Section first shows the concept of the adaptive stationary mechanism, then further enhances data reuse with a hybrid stationary scheme in a tile granularity, and finally presents the whole strategy. 

\begin{figure}
    \centering
    \includegraphics[width=1\linewidth]{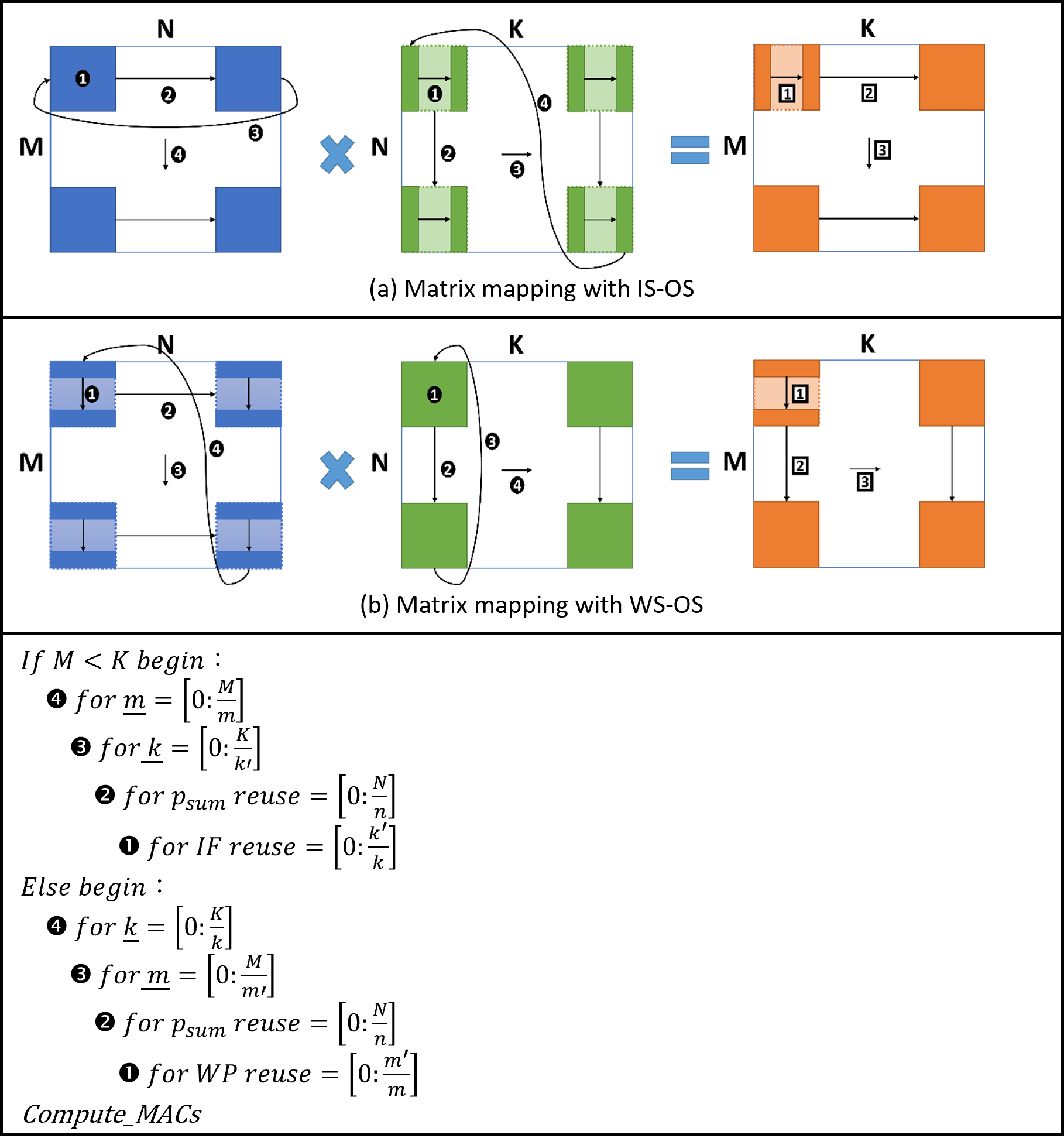}
    \caption{Matrix Mapping for Matrix-Matrix Multiplication with Proposed Stationary Schemes}
    \label{fig:PSS}
\end{figure}

\subsection{Adaptive Mechanism} 
The adaptive mechanism is constructed to optimize the selection of the most efficient stationary data scheme in a tile granularity for specific EMA needs during matrix computations. The core decision focuses on choosing between IS or WS, determined by evaluating the EMA reduction for the matrix. In current accelerators, PE arrays are often organized in a square formation like $8 \times 8$ or $16 \times 16$. This design choice facilitates efficient tile computation mapping onto the processing elements, where $m$, $n$, and $k$ are roughly equal.
For existed accelerators, each PE arrays usually consists of a squared sized PEs, such that $8 \times 8$, or $16 \times 16$. That is, for the consideration of well porting tile computation to the processing elements, $m$, $n$, and $k$ are approximately equivalent.
As demonstrated in Table~\ref{tab:ema_for_stationary_schemes}, the input matrix primarily reduces total EMA when the input stationary scheme is applied, decreasing EMA from $K \times (MN)$ to $MN$. In contrast, under the WS scheme, the weight matrix EMA is reduced by a factor of $M$. Thus, selecting either IS or WS depends simply on comparing $MN$ and $NK$, corresponding to the input and weight matrix sizes, respectively. 

The decision is governed by the expression $MN - NK = N(M-K)$. When $M < K$, the result is negative, suggesting IS is more efficient than WS. Conversely, when the result is zero or positive ($M \geq K$), WS is more advantageous and preferred. This straightforward condition highlights the minimal overhead in decision-making hardware, which merely compares the input matrix's row count with the weight matrix's column count before performing matrix multiplication. By dynamically adjusting the stationary scheme at each computing phase, we achieve notable EMA reductions compared to a static scheme. For example, in a BERT model during the linear projection stage of the query, the EMA reduction for the specified matrix can exceed 94\%, as depicted in Table~\ref{tab:ema_diff}. This input-length adaptive stationary scheme maximizes resource efficiency.

\subsection{Hybrid Strategy in a Tile Granularity}
Fig.~\ref{fig:CSS}(b) and (c) show that the internal memory capacity for partial sums would be up to $K$ and $M$ registers for the respective data-reuse strategies of IS and WS. While (d) and (e) in Fig.~\ref{fig:CSS} indicate that internal memory capacity reduces to $n$, which also is also equivalent to the shared dimension of input tile's column and weight tile's row. Consequently, to fully leverage the ability of decreasing EMA by IS or WS, and to consider the internal memory usage of partial sums for hardware implementation, we introduce the hybrid strategy based on the above adaptive mechanism. 

This strategy, in addition to the reuse of temporal data by IS or WS, incorporates a spatial reuse strategy in a tile granularity by combining the input/weight stationary scheme with the output stationary scheme. This hybrid approach not only minimizes internal memory usage, but also ensures that partial sums are not stored in internal memory until the final results are generated. Therefore, the spatial reuse of partial sums successfully prevents frequent simultaneous reading and writing of data externally. 

\subsection{Proposed Stationary Scheme}

The methodology for optimizing matrix-matrix multiplication through our innovative \textbf{tile-based adaptive stationary} approach is elucidated in Fig.~\ref{fig:PSS}. In this illustration, $Psum$ denotes partial sums, $IF$ refers to input features, and $WP$ signifies weight parameters. The choice between IS-OS and WS-OS strategies is based on the comparative dimensions of the input for linear projection computation; specifically, if $M$ is less than $K$, we select IS-WS as shown in Fig.~\ref{fig:PSS}(a). In contrast, if this condition is not met, OS-WS is used, as illustrated in Fig.~\ref{fig:PSS}(b). Within the diagram, the symbols $k'$ and $m'$ signify the number of partial sums stored internally, which is determined by the accelerator's internal memory capacity. 

For the IS-WS strategy, an input tile remains static for $k' / k$ iterations, multiplying with weight tiles placed in $k$ different positions along the $K$ dimension, shifting by a distance $k$ each cycle, indicated by \blackcircled{1}. Post temporal reuse via IS, we exploit $k'$ partial sums' spatial reuse through a row-oriented OS, shown by an arrow at \blackcircled{2}. Once the input tile has exhausted the $N$ dimension of the input matrix, it resets as denoted by \blackcircled{3}, and the weight tile shifts rightward by $k'$ to a new column set, continuing this process until the weight matrix is fully processed. Subsequently, the input tile moves downward by $m$ and the sequence repeats. 

In scenarios where $M \geq K$, the \textbf{WS-OS} strategy is activated via our adaptive scheme. Initially, the weight tile marked by \blackcircled{1} is fixed and reused for $m' / m$ iterations, with the input tile shifting downward, as indicated by an arrow with \blackcircled{1}. Upon maximizing temporal locality through WS, we aim to reuse partial sums, leveraging spatial locality. As execution proceeds, both input and weight tiles shift horizontally and vertically, respectively, as portrayed by an arrow at \blackcircled{2}, repeating the described sequence. Once the weight tile reaches the weight matrix's lower boundary, as shown by an arrow at \blackcircled{3}, it resets to the starting position while the input tile transitions to the subsequent row set, sliding anew from west to east across the input matrix. This series persists for $M / m'$ iterations until the input tile fully traverses the input matrix once and returns to the matrix's upper-left corner, marked by an arrow at \blackcircled{4}. This comprehensive cycle persists until the weight tile completely explores the weight matrix.

\section{Experimental Results}

\begin{table}[htb]
    \centering
    \caption{Comparison of EMA values for different sequence lengths in the Wav2Vec2.0-large model, where \textit{seq\_len} represents for the sequence length and \textit{ss.} is for stationary scheme}
    \label{tab:ema_diff}
    \begin{tabular}{|c|c|c|c|c|}
        \hline
        \textbf{seq\_len} & \textbf{IS} & \textbf{WS} & \textbf{IS-WS} & \textbf{optimal ss.} \\
        \hline
        115 & \num{1.18E5} & \num{1.04E6} & \num{-9.22E5} & IS \\
        384 & \num{3.93E5} & \num{1.04E6} & \num{-6.47E5} & IS \\
        1565 & \num{1.60E6} & \num{1.05E6} & \num{5.54E5} & WS \\
        15000 & \num{1.54E7} & \num{1.06E6} & \num{1.43E7} & WS \\
        \hline
    \end{tabular}

\end{table}

To show the advantages of the proposed approach, this Section shows the results evaluated on the audio and natural language processing tasks.
Table~\ref{tab:ema_diff} presents the EMA differences with the matrix applied in the stationary scheme during inference of the Wav2Vec2.0-large automatic speech recognition model \cite{baevski2020wav2vec} evaluated on the LibriSpeech \cite{panayotov2015librispeech} dataset. In this dataset, the shortest audio is about 2.3 seconds (115 tokens), and the longest one is 31.3 seconds (1565 tokens), with an average length of 7.6 seconds (384 tokens). Table~\ref{tab:ema_diff} details the EMA values for the specified matrix where data reuse is applied across these sequence lengths. We also provide EMA values for recognizing lengthy speech sequences. For sequences exceeding the maximum length, they are usually segmented into chunks for inference. Here, the inference for increasing sequence length corresponds to more rows in the input matrix, maintaining the same computation flow and EMA analysis for linear projection. Table~\ref{fig:PSS} shows that varying sequence lengths influence the optimal stationary scheme choice, indicating that \textbf{TAS}'s adaptive mechanism compensates for limitations in fixed schemes.

\begin{table}[ht]
\centering
\renewcommand{\arraystretch}{1} %
\caption{Measurement of Computing Energy Cost for BERT-BASE with Naive Implementation, AYAKA\cite{qin2024ayaka}'s Optimization, and Ours across Layers}
\label{tab:energy_comparison}
\begin{tabular}{|c|c|c|c|c|c|}
\hline
\multirow{2}{*}{\textbf{Layer ID}} & \multirow{2}{*}{\textbf{Naïve (A)}} & \multirow{2}{*}{\textbf{\cite{qin2024ayaka} (B)}} & \multirow{2}{*}{\textbf{Ours (C)}} & \multicolumn{2}{c|}{\textbf{Reduction}} \\ \cline{5-6} 
&  &  &  & \textbf{$\frac{A-B}{A}$} & \textbf{$\frac{A-C}{A}$} \\ \hline
\hline
    0  & 65.81 & 35.76 & 1.89 & 48.47\% & 97.17\% \\ \hline
    1  & 66.30 & 35.05 & 1.90 & 48.86\% & 97.15\% \\ \hline
    2  & 67.65 & 37.30 & 1.94 & 49.88\% & 97.09\% \\ \hline
    3  & 67.44 & 37.13 & 1.93 & 49.72\% & 97.10\% \\ \hline
    4  & 67.40 & 36.23 & 1.93 & 49.69\% & 97.10\% \\ \hline
    5  & 67.42 & 35.35 & 1.93 & 49.70\% & 97.10\% \\ \hline
    6  & 67.35 & 37.40 & 1.93 & 49.65\% & 97.10\% \\ \hline
    7  & 64.46 & 35.28 & 1.85 & 47.40\% & 97.23\% \\ \hline
    8  & 67.44 & 33.44 & 1.93 & 49.72\% & 97.10\% \\ \hline
    9  & 67.55 & 35.12 & 1.94 & 49.80\% & 97.09\% \\ \hline
    10 & 65.04 & 34.63 & 1.86 & 47.86\% & 97.20\% \\ \hline
    11 & 64.74 & 34.59 & 1.85 & 47.62\% & 97.21\% \\ \hline
    12 & 66.55 & 35.61 & 1.91 & 49.03\% & 97.14\% \\ \hline
\end{tabular}
\end{table}

Table~\ref{tab:energy_comparison} shows the energy consumption in the BERT-Base model by appling our method to \cite{qin2024ayaka}. The computational energy cost includes both external data transfer and internal chip processing, following the same approach and energy numbers in \cite{qin2024ayaka}. Basically, the energy consumed by external data transmission is 10 to 100 times greater than that of internal chip computation. To simplify the effective simulation of computing energy costs, measurements can be efficiently taken by evaluating the EMA ratio across various stationary schemes. The fixed stationary scheme introduced by \cite{qin2024ayaka} results in an approximate 48\% reduction in energy usage during BERT-Base model inference, on average, when compared to a basic implementation lacking stationary schemes. Our proposed stationary scheme achieves an approximately 97\% reduction in energy consumption, providing double the energy efficiency compared to the stationary scheme from \cite{qin2024ayaka}.

\section{Conclusion}

This paper introduces the data ruse strategy of \textbf{TAS}, an adaptive and efficient approach to lower EMA in transformer accelerators. TAS dynamically adjusts the stationary scheme in a tile granularity according to input sequence length, significantly cutting EMA and tackling major energy bottleneck challenges in modern transformer accelerators. Experiments reveal TAS consistently outperforms fixed schemes (IS, WS, OS) across different transformer models, reducing EMA by over 97\% in most scenarios. Additionally, TAS coordinates well with existing attention optimizations and hardware accelerators, offering a flexible, energy-saving solution for large-scale transformer models.

\bibliographystyle{IEEEtran}

\bibliography{IEEEabrv,bib/ieeeBSTcontrol,bib/thesis}

\end{document}